\documentclass[runningheads]{llncs}
\usepackage[hidelinks]{hyperref}

\usepackage[T1]{fontenc}
\usepackage{graphicx}
\usepackage{bbding}
\usepackage{subfigure}
\usepackage{enumitem}
\usepackage{amssymb}
\usepackage{amsmath}
\usepackage{multirow}
\usepackage{booktabs}
\usepackage{longtable}
\usepackage{orcidlink}
\begin{document}
\title{Ada-TransGNN: An Air Quality Prediction Model Based On Adaptive Graph Convolutional Networks}
\titlerunning{Ada-TransGNN for Air Quality Prediction}

\author{Dan Wang\textsuperscript{1}\orcidlink{0009-0004-8932-4913},
Feng Jiang\textsuperscript{2}\orcidlink{0009-0000-2878-7827}, 
Zhanquan Wang\textsuperscript{1}\Envelope}

\authorrunning{D. Wang et al.}
\institute{
\textsuperscript{1}School of Information Science and Engineering, East China University of Science and Technology, Shanghai 200237, China \\
\textsuperscript{2}School of Computer Science and Technology, Changsha University of Science and Technology, Changsha 410114 China\\
\email{zhqwang@ecust.edu.cn}
}

\maketitle              

\begin{abstract}
Accurate air quality prediction is becoming increasingly important in the environmental field. To address issues such as low prediction accuracy and slow real-time updates in existing models, which lead to lagging prediction results, we propose a Transformer-based spatiotemporal data prediction method (Ada-TransGNN) that integrates global spatial semantics and temporal behavior. The model constructs an efficient and collaborative spatiotemporal block set comprising a multi-head attention mechanism and a graph convolutional network to extract dynamically changing spatiotemporal dependency features from complex air quality monitoring data. Considering the interaction relationships between different monitoring points, we propose an adaptive graph structure learning module, which combines spatiotemporal dependency features in a data-driven manner to learn the optimal graph structure, thereby more accurately capturing the spatial relationships between monitoring points. Additionally, we design an auxiliary task learning module that enhances the decoding capability of temporal relationships by integrating spatial context information into the optimal graph structure representation, effectively improving the accuracy of prediction results. We conducted comprehensive evaluations on a benchmark dataset and a novel dataset (Mete-air). The results demonstrate that our model outperforms existing state-of-the-art prediction models in short-term and long-term predictions.
\end{abstract}
\section{Introduction}
With the rapid development of the global economy, air pollution has become a serious social issue, drawing significant attention. Studies [1,2] indicate that long-term exposure to air pollution is linked to various health problems. Air pollutants mainly include particulate matter and gases, and according to the WHO [3], air pollution causes over 7 million deaths annually. Accurately predicting air quality index (AQI) and pollutant concentrations is crucial for both public health and environmental protection.\\
However, AQI forecasting for the whole country faces several challenges. First, many factors affect air quality. Local meteorological conditions affect the concentration of particulate matter in the air[4], and there are complex interactions among these factors, making it difficult to specify the specific weight of each factor. Second, AQI shows a complex time dependence in the time dimension, with obvious periodicity and trend characteristics. As shown in Figure 1 (a), data from the same monitoring site on different dates tend to show similar patterns of change. Again, AQI also has a complex spatial dependence in the spatial dimension. Studies [5] have shown that industrial areas, traffic congestion areas, and densely populated areas significantly reduce air quality. In addition, air quality is not only affected by local pollution sources but also by neighboring areas. For example, as shown in Figure 1 (b), an area shows an increase in its pollution index due to the impact of typhoons in the surrounding areas.
\begin{figure}[h]
	\subfigure[temporal dependence]{
		\begin{minipage}[b]{0.45\linewidth}
			\centering
			\includegraphics[scale=0.29]{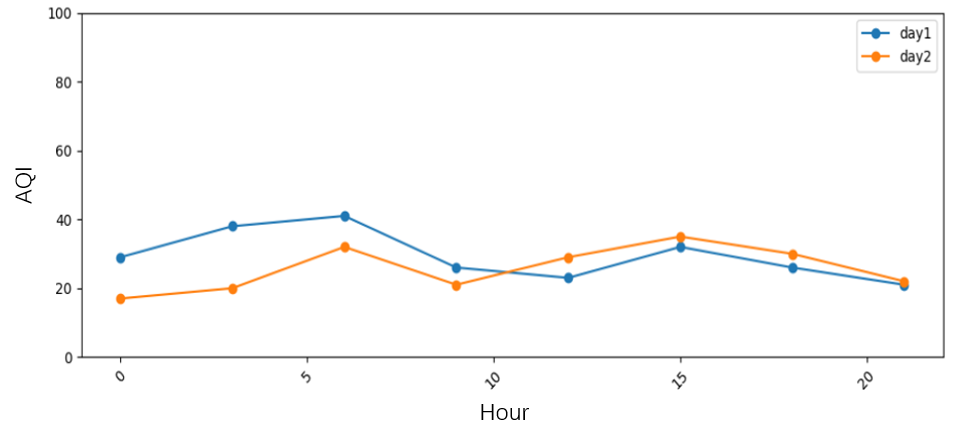} 
		\end{minipage}
	}
	\subfigure[spatial dependence]{
		\begin{minipage}[b]{0.45\linewidth}
			\centering
			\includegraphics[scale=0.31]{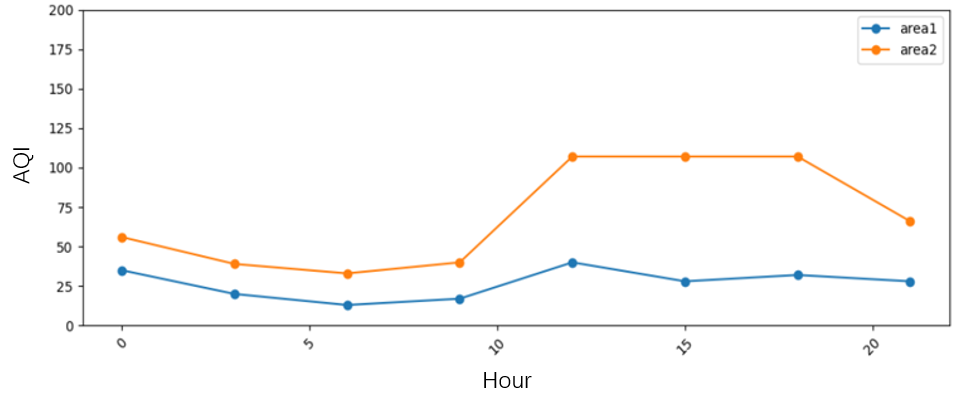}
		\end{minipage}
	}
	\caption{spatio - temporal dependences} 
	\label{fig:spatio-temporal}
\end{figure}
\\To achieve more efficient air quality prediction, existing studies have made significant progress based on graph neural networks and Transformer. However, these approaches underutilize the complex spatio-temporal correlations. For example, approaches[6-8] based STGNN typically learn spatial relationships through GNNs and capture temporal dependencies in combination with RNNs or CNNs. However, these methods mostly build graph structures based on predefined adjacency matrices, limiting the ability of the model to cope with dynamic data and complex relationships. On the other hand, Transformer-based methods [9-11] usually utilize position encoders to capture dependencies between time points and capture spatial relationships of nodes through self-attention mechanisms. However, most of the existing Transformer-based models pay more attention to the temporal dependencies of the sequences when dealing with spatio-temporal data, while ignoring the importance of the spatial locations and inter-relationships between the monitored points.\\
To solve the above problems, an adaptive spatio-temporal graph neural network (Ada-TransGNN) is proposed in this paper. First, an adaptive graph structure learning module is designed, which aims to learn the optimal graph structure guided by node attributes; second, the spatio-temporal correlation of complex data is captured by stacking multiple spatio-temporal blocks for AQI prediction, where each spatio-temporal block consists of a multiple attention mechanism and a graph convolutional network. In addition, for the same point coordinates, which may have different spatial neighbors in different training steps, auxiliary task learning is proposed to introduce Moran coefficients to capture the dynamic spatial relationships. To evaluate the effectiveness of the proposed model, we conducted experiments on two real-world datasets, and the results demonstrate the superiority of our approach concerning existing methods. Our contributions are as follows:
\begin{enumerate}[label=(\arabic*)]
	\item A novel Ada-TransGNN with adaptive augmented representation and fusion of spatio-temporal relations is proposed for AQI prediction, which effectively achieves a deeper understanding of global spatial semantics and temporal behaviors for more accurate AQI prediction.
	\item An adaptive graph learning module is proposed, which learns the optimal graph structure under the guidance of node attributes to enhance the spatial correlation of the global monitoring points, to obtain more effective spatio-temporal features. Meanwhile, Moran coefficients are introduced to quantify the spatial autocorrelation, determine the degree of aggregation among monitoring stations, and realize more effective weight allocation.
	\item Extensive experiments on an existing Chinese air quality dataset as well as a newly released dataset (mete-air) demonstrate that the model achieves better prediction performance compared with the existing baseline.
\end{enumerate}
\section{Related Works}
\subsection{Air Quality Prediction}
Recently, research methods for AQI prediction can be broadly categorized into two groups: deterministic methods and statistical methods.\\
Deterministic methods build prediction models based on physical laws and equations. CMAQ[12] is a widely used air quality prediction system that models various atmospheric processes. Studies[13]discusses the impact of meteorological parameters on prediction accuracy. However, the deterministic approach is highly sensitive to input data quality, and missing or inaccurate data can significantly reduce model accuracy. Additionally, parameterization often requires complex and time-consuming assumptions about physical and chemical processes.\\
Statistical approaches make predictions by analyzing patterns and trends in historical data, which do not strictly rely on physical and chemical models, but also show advantages in dealing with nonlinear relationships and uncertainties. These methods can be further categorized into machine learning and deep learning. However, machine learning models usually require manual selection and extraction of features, a process that relies on domain knowledge and experience, limiting the applicability and expressiveness of the models. Deep learning models often automatically learn and extract key features from data by stacking multiple neural networks. For example, RCL-Learning[14] combines residual neural networks and convolutional long and short-term memory networks to extract spatial distribution characteristics of data in multiple cities using big data association rules; HighAir[15] adopts an encoder-decoder architecture to dynamically adjust the weights according to the wind direction to capture the impact of dynamic factors on air quality. The above models focus on spatial dependency analysis within cities while ignoring the geographic location correlation among multiple cities or monitoring stations. To further capture cross-city spatial dependencies, AirFormer[16] is based on the Transformer architecture and employs a two-stage design to capture complex dependencies, and GAGNN[17] introduces a network of differentiable groups to discover potential dependencies between cities and generate city groups. Although these models integrate monitoring data from multiple sites on a national scale, the lack of effective treatment for monitoring sites with severe data missing limits their prediction accuracy and generalization ability.
\subsection{Graph Structure Learning}
Graph Neural Networks (GNN) can capture dependencies in non-Euclidean space and learn relationships between variables through a generalized graph structure. DCRNN[6] is based on the natural topology of traffic network component graphs and combines with temporal data to make predictions; ST-ResNet[18] utilizes the urban road grid's Euclidean distance as the edge weights of the graph to achieve prediction. However, these early methods either relied on natural topology or on certain metrics to build the graph, which may lead to the construction of a graph containing biases or noise, thus impairing the prediction performance. For this reason, research in recent years has shifted to optimizing the optimal structure of the graph to further improve the performance of GNNs. GraphNAS[19] is a graph neural structure search method based on reinforcement learning and takes the whole graph as learnable parameters and dynamically adjusts the graph structure by optimizing the GNN parameters. A model Graph WaveNet based on WaveNet structure[20] uses an adaptive graph convolutional layer to learn a normalized adaptive adjacency matrix and MTGNN[21]generates adaptive graphs by combining the external features of the nodes through a graph learning layer to further optimize the dependency modeling among the nodes. Unlike the above approaches, this study learns the optimal graph structure adaptively from node attributes to dynamically reflect the real associations between nodes.

\section{Ada-TransGNN}
\subsection{Problem description}
Given the historical air quality monitoring records of $N$ monitoring stations in a national region, the task of air quality forecasting is to predict future air quality conditions at each station. We first introduce some notations that will be used in this paper. According to previous studies, AQI forecasting is formulated as a time series forecasting problem. Let $X_t \in \mathbb{R}^{N \times C}$ denote the air quality index at time step $t$, where $C$ is the characteristic dimension of each node, including air pollutants and external factors. The region detected by $N$ monitoring stations and its neighboring regions is defined as a directed graph $G = (V, E, A)$, where $|V|=N$ is the set of nodes representing the monitoring stations, $E$ is a set of edges, and $A \in \mathbb{R}^{N \times N}$ is the adjacency matrix of node similarity between any pair of nodes. The objective of the AQI prediction problem is to learn a function $F(\cdot)$ that predicts the observations for the next $\tau$ steps, given the readings at all stations for the past $T$ time steps:
\begin{equation}
	(X_{t+1}, X_{t+2}, \dots, X_{t+\tau}) = F(X_{t-T+1}, X_{t-T+2}, \dots, X_t, G)
	\label{eq1}
\end{equation}
\subsection{Methods}
Figure 2 illustrates the framework of the Ada-TransGNN model, which mainly consists of a data embedding layer, an adaptive graph structure learning module, L spatio-temporal blocks, and two output layers. We will carefully describe each module in the following.
\begin{figure}[h]
	\includegraphics[scale=0.4]{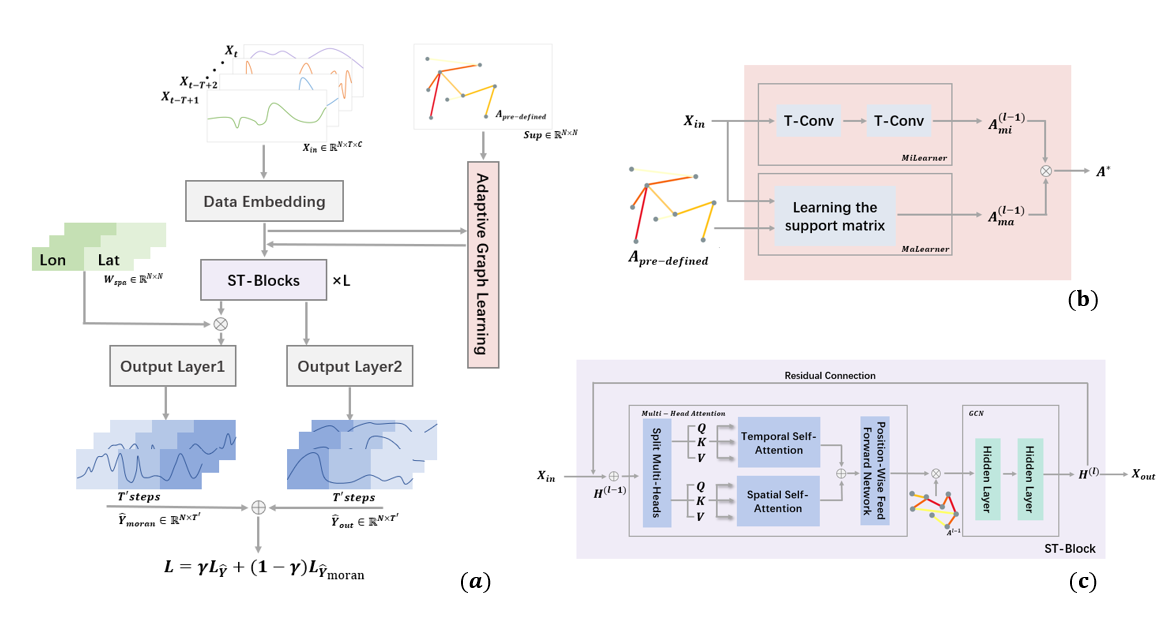}
	\centering
	\caption{The framework of Ada-TransGNN}
	\label{figure}
\end{figure}
\subsection{Adaptive Graph Structure Learning}
To solve the problem of predefined graph structure methods, as shown in Figure (2)b, we propose an adaptive graph structure learning component to obtain the optimal graph structure in a data-driven approach. The component is mainly divided into two modules, a macro learning module, which focuses primarily on the learning and optimization of the overall graph structure, and a micro-learning module, which mainly focuses on the learning and optimization of the local node features and neighborhood relations.

\subsubsection{Macro-Learning Module}
From a macroscopic point of view, the relationships of nodes in a graph are usually considered relatively stable. In contrast, predefined graph structures are typically constructed based on simple geometric properties such as Euclidean spatial distances[22], which can only represent relatively singular properties, such as the proximity of physical distances.\\
To completely capture the correlation between all monitoring points, we propose a macro learning module. This module dynamically learns and adjusts the spatial relationships among nodes based on input node attributes, including the concentration of multiple pollutants. Specifically, the neighbor matrix $A$ generated by the macro learning module is shown in Eqs.~\eqref{eq2},~\eqref{eq3}:
\begin{equation}
	A^{(\ell)} = f(\{x_i\}) \label{eq2}
\end{equation}
\begin{equation}
	A_{ma}=A^{(l)} + A^0 \label{eq3}
\end{equation}
where $x_i\in\mathbb{R}^{N\times D}$, $N$ is the number of monitoring points and $D$ is the dimension of the node attributes; $A^0$ denotes the initial space-based adjacency matrix.
\subsubsection{Micro-learning module}
From a microscopic point of view, the relationship of nodes is usually affected by some short-term events, such as morning and evening peaks, weather changes and other factors. The occurrence of these events tends to be trending and cyclical.To completely capture the complex micro-variation relationships between monitoring points, we propose a micro-learning module. The module extracts and learns the input detection point features through two convolutional neural networks CNN and activation function ReLU to capture the node state changes between different time points. Specifically, the neighbor matrix A generated by the micro-learning module is shown in Eq.~\eqref{eq4}:
\begin{equation}
	A_{mi}=\text{Conv2d}(\text{ReLU}(\text{Conv2d}(x_i)))\label{eq4}
\end{equation}
where \(x_i \in \mathbb{R}^{N\times T\times D}\), where \(T\) is the time step.\\
With the above two modules, we can obtain the graph adjacency matrix \(A_{ma}\) at the macro level and \(A_{mi}\) at the micro level respectively. To go further to obtain the optimal graph adjacency matrix, we will use an activation layer and a subsumption layer to process these adjacencies according to the matrices to come out with the optimal graph adjacency matrix as shown in Eq.~\eqref{eq5}:
\begin{equation}
	A^{(1)}=\text{Norm}(\text{ReLU}(A_{ma} \cdot A_{mi}))\label{eq5}
\end{equation}
\subsection{ST-Block}
To capture global node features, i.e., features between each node and all other nodes, Ada-TransGNN proposes to capture complex spatio-temporal dependencies by stacking spatio-temporal blocks. As in Figure 2(c), each spatio-temporal block has two parts. The first one is a multi-attention mechanism layer and the second one is a graph convolution layer, and residual connections are added between each spatio-temporal block to better train the model.
\subsubsection{Multi-attention mechanism}
Dependencies exist between air quality conditions within different time segments, and the dependencies keep changing depending on the situation. Therefore, we employ a multi-head temporal attention module to discover dependencies in time series. The canonical self-attention [23] is defined based on tuples, and the scaling click operation is performed based on the relationship between key and query to scale the value. As shown in Eq.~\eqref{eq6}:
\begin{equation}
	\text{Attention}(Q, K, V)=\text{softmax}\left(\frac{QK^{T}}{\sqrt{d_{k}}}\right)V\label{eq6}
\end{equation}
where \(Q\) is the query, \(K\) is the key, \(V\) is the value, and \(d_{k}\) is the dimension of the query and key.\\
In order to further enhance the model fitting ability, Ada-TransGNN employs a multi-head attention mechanism, as shown in Eqs.~\eqref{eq7},~\eqref{eq8}:
\begin{equation}
	\text{head}_i = \text{Attention}(Q W_i^Q, K W_i^K, V W_i^V)\label{eq7}
\end{equation}
\begin{equation}
	\text{Attention}(Q_i, K_i, V_i) = \text{softmax}\left(\frac{Q_i K_i^T}{\sqrt{d_k}}\right) V_i\label{eq8}
\end{equation}
where \(i\) denotes the \(i\)th attention head and \(d_k = d_v = \frac{d_{\text{model}}}{i}\). Each attention head has its own weight matrix and computes independent attention scores and outputs. Ada-TransGNN, by using multiple attention heads, can learn multiple different attention representations in parallel, capturing the dynamic temporal trends at different levels in the sequence.
\subsubsection{Graph Convolution}
Since monitoring stations are generally placed irregularly in different areas and may be sparsely distributed in the same area. Therefore, it is difficult to handle the task with conventional GCN.\\
According to studies [24,25], graph convolution can be categorized into two types, one is spectrum-based and the other is domain-based. In this paper, we use domain-based Chebyshev graph convolution [26], which centers on the use of Chebyshev polynomials to approximate the convolution operation on a graph, given a graph \(G=(V, E)\), where \(V\) is the set of nodes, \(E\) is the set of edges, and the feature of each node \(v_i\) is denoted as \(x_i\), the specific formula is shown in Eq.~\eqref{eq9}:
\begin{equation}
	\text{ChebConv}(X, A)=\sum_{k = 0}^{K - 1} \theta_k T_k(\tilde{L})X\Theta_k\label{eq9}
\end{equation}
where \(X\) is the input feature matrix and each row corresponds to the eigenvector of a node; \(A\) is the adjacency matrix; \(\tilde{L} = \frac{2L}{\gamma_{\max}} - I\) is the Chebyshev approximation of the normalized Laplacian matrix, where \(L = D - A\) is the Laplacian matrix, \(D\) is the degree matrix, and \(\gamma_{\max}\) is the maximal eigenvalue of the Laplacian matrix; \(T_k\) is the \(k\)th - order Chebyshev polynomial; and \(\theta_k\) and \(\Theta_k\) are learnable parameters.
\subsection{Assisted Task Learning}
Geographic data of monitoring points, such as latitude and longitude information, are usually considered in AQI prediction tasks, and the monitoring data are usually related to their geography in some relationship. Moran coefficient[27] is mainly used to measure the strength of spatial connection between monitoring points, and in this task, we introduce the Moran coefficient to help the model better understand the spatial structure and relationship of air quality data. Specifically, the spatial autocorrelation measure formula for the feature variable \(x_i\) is shown in Eqs.~\eqref{eq10},~\eqref{eq11}:
\begin{equation}
	M_i = \frac{(N - 1) \cdot z_{ij} \cdot z_{il}}{\text{den}_i}\label{eq10}
\end{equation}
\begin{equation}
	\text{den}_i =\sum_{j = 1}^{N} (\sum_{j = 1}^{N} w_{ij} \cdot z_{ij})^2 \label{eq11}
\end{equation}
where \(N\) is the total number of monitoring points and \(w_{ij}\) is the spatial weight between monitoring points \(i\) and \(j\).\\
Researches[28,29] invoked uncertainty to weight the loss function to optimize the scene geometry and semantic auxiliary tasks, and auxiliary task learning to enhance the accuracy and robustness of the model's prediction, respectively. Based on this, we propose an auxiliary task learning based on the Moran coefficients by calculating the Moran coefficients \(M_i\) and using them as an auxiliary loss function to optimize the prediction performance of the main task.
\subsection{Prediction and Optimization}
We calculate the spatial weight \( y_{\text{moran}} \) of all monitoring point eigenvalues \(x_i\) according to Eq.~\eqref{eq9} in each training step, and Ada-TransGNN creates two output layers for prediction. Meanwhile, considering the error accumulation and model efficiency, we choose long-term prediction by directly converting the hidden dimensions of the output layers to the desired dimensions, as shown in Eqs.~\eqref{eq12},~\eqref{eq13}:
\begin{equation}
	\hat{y} = \text{Conv1d}(\text{Conv2d}(\text{ReLU}(\text{Conv2d}(x_{\text{hid}}))))\label{eq12}
\end{equation}
\begin{equation}
	\hat{y}_{\text{moran}} = \text{Linear}(\text{ReLU}(\text{Linear}(y_{\text{moran}})))\label{eq13}
\end{equation}
where \(\hat{y}, \hat{y}_{\text{moran}} \in \mathbb{R}^{N\times T'\times C}\) is the prediction result of \(T'\) step.\\
Finally, we get the primary task prediction value \(\hat{y}\) and auxiliary task prediction value \(\hat{y}_{\text{moran}}\). We choose the mean square error MSE for the loss Loss calculation and update the model parameters according to stochastic gradient descent. Considering the task trade-off problem in multi-task learning (MTL), the uncertainty weighting method proposed by Kendall et al. [19] is borrowed to dynamically adjust the contribution weights of the main and auxiliary tasks.The specific formula is shown in Eq.~\eqref{eq15}:
\begin{equation}
	\text{Loss}=\lambda\cdot\text{MSE}(\hat{y},y)+(1 - \lambda)\cdot\text{MSE}(\hat{y}_{\text{moran}},y_{\text{moran}})\label{eq15}
\end{equation}
where \(\lambda\) denotes the auxiliary task learning weight. In traditional methods, \(\lambda\) is usually determined by experimental tuning, but inspired by the idea of multi - task uncertainty weighting, we can let the model learn the optimal \(\lambda\) automatically, i.e., Eq.~\eqref{eq16}:
\begin{equation}
	\lambda=\frac{1}{2\sigma^{2}}\label{eq16}
\end{equation}
where \(\sigma^{2}\) represents the task uncertainty, and higher uncertainty means less task weight and vice versa. \\
In this way, we ensure the accuracy of the main task while fully utilizing the spatial dependence information provided by the auxiliary tasks to enhance the overall prediction effect.
\section{Experiments}
This section evaluates the effectiveness of the proposed Ada-TransGNN model by conducting extensive experiments on two real-world datasets. For reproducibility, we provide the code and datasets on Github\footnote{https://github.com/wwwangddd/Ada-TransGNN}.And in this task, two evaluation metrics are used to measure the performance of different models, including Mean Absolute Error (MAE), and Root Mean Square Error (RMSE).
\subsection{Datasets}
We evaluate the performance of the Ada-TransGNN model on two Chinese city air quality datasets, described as follows: 
\begin{enumerate}[label=(\arabic*)]
	\item Dataset 1 [17]: this dataset contains AQI data and weather data collected between January 1, 2017, and April 30, 2019, for 209 cities. Cities with serious missing data were excluded, and 20,370 valid samples were finally retained.
	\item Dataset 2 (mete - air): Each record contains six pollutant concentrations and four meteorological data. We divided the dataset according to the ratio of 7:1:2, 70\% for the training set, 10\% for the validation set, and 20\% for the test set. The specific information of the dataset is shown in Table 1.\\
	According to the existing studies [17,31], we designed different prediction tasks on the two datasets, predicting the AQI for the next 6 time steps based on the data of the past 24 time steps (1 hour each) on dataset 1, and predicting the AQI for the next 24 time steps (72 hours) based on the data of the past 24 time steps (3 hours each) on dataset 2.
\end{enumerate}
\begin{table}
	\caption{Detailed Information about Mete - Air}
	\centering
	\begin{tabular}{|l|p{8cm}|}
		\hline
		Name & Mete - Air \\
		\hline
		Start - time & 2020 - 01 - 01 \\
		\hline
		End - time & 2024 - 02 - 21 \\
		\hline
		Time interval & 3 Hours \\
		\hline
		Numbers of stations & 120 \\
		\hline
		Timesteps & 36312 \\
		\hline
		Features & PM2.5, PM10, NO2, CO, O3, SO2, Air Temperature,\\
		& Wind Direction,Wind Speed, Rainfall \\
		\hline
	\end{tabular}
\end{table}

\subsection{Baselines Methods}
To validate the performance of Ada-TransGNN, we compared it with existing baseline models, and the selected baseline models can be categorized into the following four groups: (1) traditional time series models (ARIMA, LSTM); (2) models based on graph neural networks (STGCN, DMSTGCN, FCSTGNN); (3) models based on the attention mechanism (GMAN, ASTGCN, STAGCN); (4) air quality prediction model (GAGNN).\\
ARIMA: predicts values at \(q\) future time points based on \(p\) past time points.\\
LSTM [31]: A type of RNN that predicts \(q\) future time points using gating mechanisms and memory units.\\
STGCN(2018) [32]:Uses spatial spectrogram convolution and temporal TCN to form a spatio-temporal convolution block.\\
DMSTGCN(2021) [33]: Combines dynamic graph construction, residual networks, and graph convolution for better dependency modeling.\\
FCSTGNN(2024) [34]: Models spatio-temporal dependencies using decaying graph construction and moving pool GNN layer.\\
GMAN(2020) [35]: Uses an encoder-decoder architecture with an attention-switching module to model time-step relationships.\\
ASTGCN(2019) [36]: Combines spatio-temporal attention and convolution to capture features and produce predictions.\\
STAGCN(2022) [37]: Models static and dynamic maps separately to capture global spatial and local dynamic changes.\\
GAGNN(2021) [17]: Groups cities into clusters and captures dependencies within them via a mapping matrix.
\subsection{Settings}
The experiments are implemented in the NVIDIA GeForce RTX 3090 environment using the Pytorh2.1.0 framework. The RMSprop optimizer was used for training, the batch\_size was set to 64, the number of attention heads was set to 2, and the learning rate and decay rate were 0.0001 and 0.0005, respectively.

\begin{table}[htbp]
  \centering
  \caption{Results on Dataset 1.}
  { 
    \setlength{\tabcolsep}{1pt} 
    \scriptsize          
    \begin{tabular}{p{1.8cm} *{12}{p{0.8cm}}}
      \toprule
      \multirow{2}{*}{Model} & \multicolumn{2}{c}{1h} & \multicolumn{2}{c}{2h} & \multicolumn{2}{c}{3h} & \multicolumn{2}{c}{4h} & \multicolumn{2}{c}{5h} & \multicolumn{2}{c}{6h} \\
      \cmidrule(lr){2-3} \cmidrule(lr){4-5} \cmidrule(lr){6-7} \cmidrule(lr){8-9} \cmidrule(lr){10-11} \cmidrule(lr){12-13}
      & MAE & RMSE & MAE & RMSE & MAE & RMSE & MAE & RMSE & MAE & RMSE & MAE & RMSE \\
      \midrule
      ARIMA & 17.86 & 25.29 & 18.55 & 25.33 & 18.64 & 25.43 & 18.74 & 25.57 & 18.87 & 26.74 & 18.97 & 29.96 \\
      LSTM & 22.19 & 45.89 & 23.21 & 46.21 & 24.07 & 46.57 & 24.87 & 46.95 & 25.66 & 47.37 & 25.99 & 47.27 \\
      DMSTGCN & 9.57 & 16.86 & 10.49 & 18.53 & 12.74 & 22.86 & 14.61 & 25.02 & 16.05 & 27.03 & 17.24 & \underline{29.07} \\
      STGCN & 9.32 & 16.74 & 10.45 & 18.68 & \underline{12.02} & \underline{21.94} & 14.29 & 24.52 & \underline{15.70} & \underline{26.73} & 17.20 & 29.31 \\
      FCSTGNN & 6.02 & 11.89 & 9.89 & 18.49 & 12.64 & 25.36 & \underline{14.05} & \underline{24.42} & 16.16 & 27.39 & 17.41 & 29.97 \\
      GMAN & 13.62 & 25.47 & 14.84 & 26.92 & 16.07 & 28.42 & 17.21 & 29.83 & 18.25 & 31.12 & 19.18 & 32.28 \\
      ASTGCN & 6.91 & 12.81 & 9.99 & 18.02 & 12.58 & 22.87 & 14.67 & 24.97 & 16.34 & 27.58 & 17.67 & 29.98 \\
      STAGCN & 8.18 & 14.01 & 11.01 & 19.18 & 13.39 & 23.09 & 15.33 & 26.07 & 16.89 & 28.37 & 18.16 & 30.16 \\
      GAGNN & \underline{5.83} & \underline{11.38} & \underline{9.42} & \underline{17.26} & 12.04 & 24.19 & 14.56 & 25.01 & 15.78 & 26.89 & \underline{17.04} & 29.12 \\
      Ours & \textbf{5.75} & \textbf{11.15} & \textbf{9.26} & \textbf{17.02} & \textbf{11.88} & \textbf{21.12} & \textbf{13.95} & \textbf{24.16} & \textbf{15.60} & \textbf{26.59} & \textbf{16.74} & \textbf{28.47} \\
      \bottomrule
    \end{tabular}
  } 
\end{table}

\begin{table}[!h]
  \centering
  \caption{Results on Dataset 2.}
  \begin{tabular}{l*{6}{p{1.6cm}}}
    \toprule
    Model & MAE & RMSE & Model & MAE & RMSE\\
    \midrule
    ARIMA & 17.52 & 22.36 & LSTM & 16.04 & 20.85\\
    DMSTGCN & 12.86 & 18.93 & STGCN & 13.05 & 19.02\\
    GMAN & 16.02 & 19.89 & ASTGCN & 17.93 & 22.83\\
    STAGCN & 15.05 & 18.74 & GAGNN & 18.05 & 23.15\\
    FCSTGNN & 12.34 & 17.29 & Ours & \textbf{11.60} & \textbf{16.04}\\
    \bottomrule
  \end{tabular}
\end{table}

\subsection{Experimental results and analysis}
We compare the proposed Ada-TransGNN model with the baseline method on two datasets and the experimental results are shown in Tables 2 and 3. Table 2 shows the cumulative results of prediction errors for each time step for the next 6 hours, and Table 3 shows the average error results for the long-term prediction for the next 72 hours. The results show that the Ada-TransGNN model outperforms the baseline model in both MAE and RMSE metrics, and performs best in both short-term and long-term forecasts. In addition, the traditional time series approach performs poorly, suggesting its inability to capture complex spatio-temporal dependencies.GAGNN performs second only to our model in dataset 1, which may be due to its reliance on graph preprocessing and a priori knowledge. FCSTGNN, ASTGCN and STGCN perform well overall, and even though the two models were initially used for traffic flow prediction, they show good air quality prediction task adaptability and generalizability. Finally, the Ada-TransGNN model also outperforms ASTGCN, suggesting that in addition to the adaptive graph structure learning module, the auxiliary task learning module also improves the model's prediction performance. On dataset 2, most of the models used for short-term prediction of traffic flow are not suitable for long-term air quality prediction, and the modes have higher errors in 1-72 hours, with FCSTGNN performing second only to our model overall.

\subsection{Ablation experiment}
We conducted an ablation study to evaluate the effectiveness of the proposed module by designing the following three variants:(1) Ada-TransGNN without the auxiliary task of learning the Moran coefficients. In this variant, the geographic data of the monitoring points are not fused to capture spatial autocorrelation; (2) Ada-TransGNN without the micro-learning module. In this variant, GCN is not used to capture instantaneous changes of nodes; (3) Ada-TransGNN without macro learning module, in this variant, predefined static graphs are directly used to capture spatio-temporal dependencies.
The results of the ablation study are shown in Figure 3, indicating that the design of the key modules is conducive to improving the predictive performance of the model. Based on the results, we conclude that (1) the performance of Ada-TransGNN is significantly improved compared to the model without Moran coefficients, indicating the importance of Moran coefficients in capturing spatial autocorrelation, and (2) the macro-learning module and the micro-learning module in the adaptive graph structure learning module further enhance the model by learning the latent attributes of the input nodes to enhance the prediction performance.
\begin{figure}[h]
	\centering
	\includegraphics[scale=0.48]{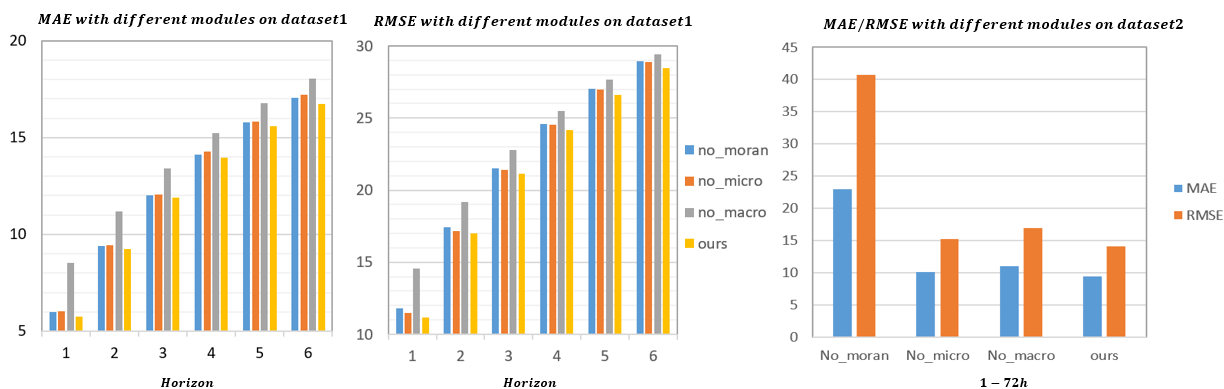}
	\caption{Ablation research}
	\label{figure}
\end{figure}
\subsection{Parametric Sensitivity experiment}
We conducted an experiment to analyze the impact of three key hyperparameters: the number of heads of the multi-head attention mechanism $h$, the number of spatio-temporal blocks $b$, and the weights of the two output layers in the loss function $\lambda$. In Figure 4, we show the percentage of the model's degradation compared to the optimal performance in the mete-air dataset 2, when varying one parameter while leaving the others at their default values. The results show that adjusting the number of heads $h$ (set to 1, 2, 5, and 10, respectively) may lead to an increase in computation leading to a decrease in performance; also, increasing the number of spatio-temporal blocks (set to 1, 3, 5, and 6, respectively) lead to more noise and overfitting, which may also lead to a decrease in performance; lastly, we found that when $\lambda$ was set to 0.5 (set to 0.1, 0.3, 0.5, and 0.7), the best performance is obtained.
\begin{figure}
	\centering
	\includegraphics[scale=0.5]{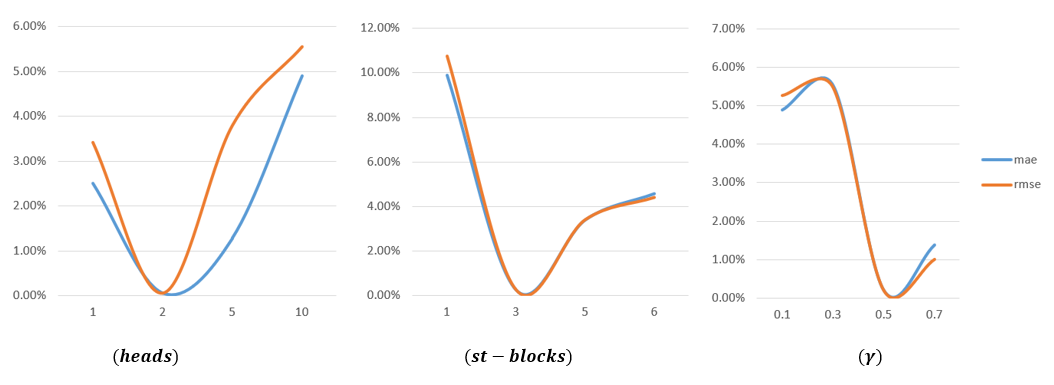}
	\caption{Parametric Sensitivity research}
	\label{figure}
\end{figure}

\section{Conclusion}
In this paper, an air quality index prediction model for Ada-TransGNN is proposed. To effectively model the spatio-temporal dependence of complex air quality data, the framework is modeled by stacking spatio-temporal blocks; to optimize the spatial relationship between monitoring points, an adaptive graph structure learning module is designed to learn the optimal graph structure through node attributes; and to capture spatial autocorrelation, an auxiliary task learning module is introduced, which is achieved by calculating Moran coefficients of nodes. In addition to using a publicly available dataset, we generated an air quality dataset called mete-air from a publicly available data site. We conducted extensive experimental evaluations on both datasets, and the results demonstrate the superiority of our approach. In the future, we will further investigate the interpretability of the model and consider introducing more domain knowledge and multi-scale features to enhance the predictive performance of the model.

%
%
%

%

\end{document}